\documentclass[a4paper, 10pt, conference]{ieeeconf}      

\usepackage[utf8x]{inputenc}

\IEEEoverridecommandlockouts                              

\usepackage{graphicx} 
\usepackage{amsmath} 
\usepackage{amssymb}  

\usepackage{bbm}
\usepackage{siunitx} 
\usepackage[breaklinks]{hyperref}
\usepackage{subcaption}
\usepackage{multirow}
\newcommand{\citegeolife}[0]{\cite{zheng2009mining,zheng2008understanding,zheng2010geolife}}

\title{\LARGE \bf
DeepStay: Stay Region Extraction from Location Trajectories using Weak Supervision
}

\author{Christian Löwens$^{1}$, Daniela Thyssens$^{1}$, Emma Andersson$^{2}$, Christina Jenkins$^{2}$, and Lars Schmidt-Thieme$^{1}$
\thanks{$^{1}$Christian Löwens, Daniela Thyssens, and Lars Schmidt-Thieme are with the Information Systems and Machine Learning Lab (ISMLL),
        University of Hildesheim, 31141 Hildesheim, Germany.
        {\tt\small loewensc@uni-hildesheim.de, \{thyssens, schmidt-thieme\}@ismll.uni-hildesheim.de}}%
\thanks{$^{2}$Emma Andersson and Christina Jenkins are with Devoteam Creative Tech, 211 20 Malmö, Sweden.
        {\tt\small \{emma.andersson, christina.jenkins\}@devoteam.com}}%
}

\begin{document}

\maketitle
\thispagestyle{empty}
\pagestyle{empty}

\begin{abstract}

Nowadays, mobile devices enable constant tracking of the user's position and location trajectories can be used to infer personal points of interest (POIs) like homes, workplaces, or stores. A common way to extract POIs is to first identify spatio-temporal regions where a user spends a significant amount of time, known as stay regions (SRs).

Common approaches to SR extraction are evaluated either solely unsupervised or on a small-scale private dataset, as popular public datasets are unlabeled. Most of these methods rely on hand-crafted features or thresholds and do not learn beyond hyperparameter optimization. Therefore, we propose a weakly and self-supervised transformer-based model called DeepStay, which is trained on location trajectories to predict stay regions. To the best of our knowledge, this is the first approach based on deep learning and the first approach that is evaluated on a public, labeled dataset. Our SR extraction method outperforms state-of-the-art methods. In addition, we conducted a limited experiment on the task of transportation mode detection from GPS trajectories using the same architecture and achieved significantly higher scores than the state-of-the-art.
Our code is available at \href{https://github.com/christianll9/deepstay}{https://github.com/christianll9/deepstay}.

\end{abstract}

\section{Introduction}

Extracting stay regions (SR) from location trajectories identifies segments where a subject stays in the same place. It supports fine-grained spatio-temporal analysis of human and animal behavior and is often an intermediate step in point of interest (POI) mapping or POI extraction.

Common SR extraction approaches apply unsupervised clustering algorithms and use thresholds for time, distance, and velocity, among others. These thresholds are determined by a qualitative analysis or a quantitative hyperparameter optimization. Typically, all experiments are performed either on small private datasets, in some cases with manually annotated labels, or on large datasets without any labels. This makes it difficult to compare different approaches and makes the problem less suited for supervised learning that requires a large amount of labeled data.

\begin{figure}
    \centering
    \includegraphics[width=\linewidth]{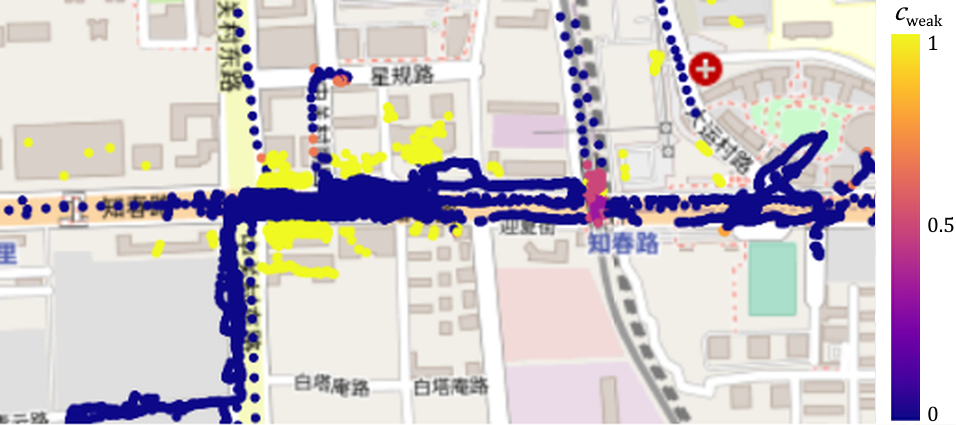}
    \caption{Weak supervision of trajectories using \href{https://www.openstreetmap.org/copyright}{OSM} data and additional heuristics. Blue points indicate weakly labeled non-stays, yellow indicate stays.}
    \label{fig:intro}
\end{figure}

Even though most trajectories do not contain ground truth SR labels, it is still possible to derive so-called "weak labels" from OpenStreetMap (OSM). For example, we can classify any location point lying within a building as part of a stay and a point near a road as part of a "non-stay" (see Figure \ref{fig:intro}). Given the large number of weak labels available, we assume that this data contains enough signal to learn useful latent representations. To this end, we apply a transformer model \cite{vaswani2017attention} that takes a trajectory as a time series of location points and classifies each point as either part of a stay or a non-stay.

To our knowledge, this is the first approach to extract SRs from trajectory data using deep learning. Furthermore, we use publicly available data for training and evaluation to ensure reproducibility. We derived a ground truth dataset from the field of activity recognition and use it to compare our model with baselines from related work.

\section{Problem Statement}
\label{sec:intro/problem/definition}
We define a location trajectory $\mathcal{X}=\{g_{1}, g_{2}, \ldots, g_{|\mathcal{X}|}\}$ as a time series of consecutive location points $g_{i}=\left(t_{i}, x_{i}, y_{i}\right)$, where $x, y\in\mathbb{R}$ denote the 2D coordinates and $t\in\mathbb{R}^{\ge0}$ the ascending timestamp. The sample rate $\Delta t_{i}=t_{i}-t_{i-1}$ is defined as the time difference between two consecutive points and is either constant or fluctuating, depending on the dataset.

\label{sec:intro/problem/definition/stay}
SR extraction can be viewed as a time series segmentation task, where the trajectory $\mathcal{X}$ is split in a set of segments $\mathcal{TS}=\{ts_1,\dots,ts_q\}$. Each segment $ts_j=(t_{{\mathrm{start}}_j}, t_{{\mathrm{end}}_j}, c_j)$ is defined by its start $t_{\mathrm{start}}$ and end time $t_{\mathrm{end}}$ and the binary class $c\in\{0,1\}$ indicating whether the user is staying at one place ($c=1$) or is moving around ($c=0$) within the time window $t_{\mathrm{start}}\le t<t_{\mathrm{end}}$. Moreover, we define
\begin{align*}
  \qquad \qquad \qquad\qquad t_{{\mathrm{start}}_1} &= t_1,\\
  \qquad \qquad \qquad\qquad t_{{\mathrm{end}}_q} &= \infty,\\
  \qquad \qquad \qquad\qquad t_{{\mathrm{start}}_j} &< t_{{\mathrm{end}}_{j}}&\forall j\in\{1,\dots,q\},\\
  \qquad \qquad \qquad\qquad t_{{\mathrm{end}}_j} &= t_{{\mathrm{start}}_{j+1}}&\forall j\in\{1,\dots,q-1\},\\
  \qquad \qquad \qquad\qquad c_j &\ne c_{j+1}&\forall j\in\{1,\dots,q-1\}.\\
\end{align*}
The set of stay regions $\mathcal{SR}$ is a subset of all segments, where
\begin{equation}
  \mathcal{SR}=\{ts_j|ts_j\in\mathcal{TS}\wedge c_j=1\}.
\end{equation}
The task of SR extraction is now to predict $\mathcal{SR}$ (and therefore $\mathcal{TS}$) solely from the trajectory data $\mathcal{X}$.

\section{Related Work}
\label{sec:related}
Trajectory segmentation is an important research topic with many examples such as activity recognition, transportation mode detection (TMD) and SR extraction. In TMD, each segment is assigned to a mode, e.g. walking, car, bus, etc. \cite{zheng2010understanding}. A special binary case of this task is SR extraction with only two possible modes: stay and non-stay. In most cases, it functions as a preprocessing step for tasks such as POI map\-ping\slash extraction\slash pre\-dic\-tion. In POI mapping, each SR is assigned to a visit to one of several POIs \cite{suzuki2019personalized}.

SR extraction identifies segments of a user's trajectory where the subject remains at the same place. A virtual location, usually the centroid of an SR, is called a stay point. So this task is also called stay point extraction\slash reco\-gni\-tion\slash iden\-ti\-fi\-ca\-tion\slash de\-tection.

\subsection{Threshold-based Clustering}
The vast majority of published work uses threshold-based spatio-temporal clustering methods, where the clusters represent stay segments of the trajectory. Commonly used thresholds are a minimum duration $T_{\mathrm{min}}$ and a maximum distance $D_{\mathrm{max}}$ \cite{kang2005extracting,li2008mining,zheng2009mining,andrade2020discovering,umair2014discovering}. Here, the task is to find the maximum sets of consecutive location points $\mathcal{P}=\{g_m,g_{m+1},\dots,g_n\}$ in the trajectory $\mathcal{X}$, such that:
\begin{align}
  \label{eq:t_tresh}
  t_{n} - t_{m} &\ge T_{\mathrm{min}}\\
  \qquad\qquad
  \label{eq:d_tresh}
  \mathrm{dist}(g_{i},g_{j}) &\le D_{\mathrm{max}}\quad&&\forall\quad g_{i}, g_{j}\in \mathcal{P}&
\end{align}
Others apply additional thresholds for velocity, acceleration, and heading change \cite{bhattacharya2012extracting,pavan2015finding,thomason2016identifying,zheng2008understanding,pavan2017mining}.

\subsection{Adapted Density-based Clustering}
Other approaches adapt density-based clustering methods such as DBSCAN \cite{ester1996density} and OPTICS \cite{ankerst1999optics}. If the trajectory is sampled at a constant rate, prolonged stays will result in dense spatial data and thus can be detected. Unlike k-means, they do not require a predetermined number of clusters, which is crucial for SR extraction.

These approaches define SR extraction more as spatial clustering rather than time series segmentation. Therefore, the constraint that the clustered points must be consecutive is not always enforced. Many extensions have been proposed to utilize the temporal information as well \cite{cbsmot,nishida2015extracting}.

\subsection{Others}
\label{sec:related/stay/others}%
The authors in \cite{ashbrook2003using} and \cite{cao2010mining} classify single location points as stay points when a GPS connection loss is detected. The algorithm proposed by \cite{stylianou2017stay} extracts SRs by searching for local minima of speed and zero crossings in acceleration within the trajectory.

\section{Methodology}
\label{sec:method}

\subsection{Architectural Overview}
\label{sec:method/architect}
\begin{figure}
    \centering
    \includegraphics[width=\linewidth]{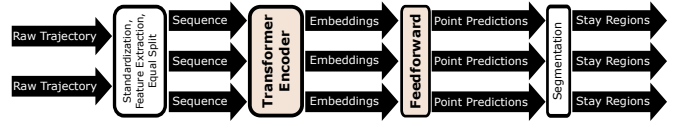}
    \caption{Overall architecture of DeepStay. Light brown colored boxes indicate trainable models.}
    \label{fig:method/architect}
\end{figure}

Figure \ref{fig:method/architect} shows the overall architecture of our model DeepStay and the intermediate results of the processing pipelines.
First, the raw trajectories are standardized and split into sequences of equal size. Furthermore, additional features are extracted to improve the performance of the subsequent transformer encoder. 
This encoder receives a sequence of constant length and outputs an embedding vector for each point comprising latent features about the point within its sequence. The following feedforward layer acts as a decoder and predicts a probability for each vector to be part of a stay.
In the next step, all consecutive points with a predicted probability above a certain threshold are grouped as SRs.

\subsubsection{Preprocessing of raw GNSS trajectories}
\label{sec:method/sp-class/preprocess}
All datasets in this work contain GNSS coordinates, such as GPS. In the first step, we project all coordinates into a 2D Cartesian system $(x,y)$ using an appropriate UTM zone \cite{langley1998utm}.

Since trajectories may have varying sample rates, we use the time difference $\Delta t$ between each point and its predecessor as an additional feature. Our preliminary experiments indicate that this approach leads to better results than using linear interpolation as proposed by \cite{moreau2021devil}.
We also add the current velocity $v$ as the ratio of the Euclidean distance and $\Delta t$ between two consecutive points as another feature.

All trajectories are chunked into sequences of equal length $n=256$. This allows the transformer encoder to be trained with multiple sequences in a single batch of size $B=64$.

Furthermore, we standardize the features $\Delta t$ and $v$ separately based on their distribution in the training set to obtain a mean of 0 and a standard deviation of 1. The standardization of the location features $x$ and $y$ is done jointly. Each sequence is subtracted from its mean $(\overline{x}_{\mathrm{seq}},\overline{y}_{\mathrm{seq}})$ and divided by the common standard deviation of the entire training set $\sigma_{x,y_{\mathrm{train}}}$ to prevent the model from memorizing specific regions. To further reduce overfitting, we rotate every sequence uniformly at random with respect to its origin $(0,0)$ before feeding it to the model. The final features of the $i$-th data point in sequence $seq$ are shown in \ref{eq:method/sp-class/preprocess/feat}.

\begin{equation}
    \label{eq:method/sp-class/preprocess/feat}
    seq_{i} =
    \left[\begin{array}{cccc}
        x_i & y_i & \Delta t_i & v_i
        \end{array}\right],\quad~seq\in\mathbb{R}^{n\times 4}
\end{equation}

\subsubsection{Transformer Encoder}
\label{sec:method/sp-class/embeddings}
We choose the encoder of the transformer model \cite{vaswani2017attention} to learn latent embeddings $emb_{i}$ for each sequence point $seq_{i}$. This allows us to predict the class probabilities pointwise instead of segmentwise. Thus, by design, segmentation and classification are performed jointly. We stick with the original setting of the base encoder including the projection and positional encoding as described in \cite{vaswani2017attention} to get the final embeddings $emb\in\mathbb{R}^{n\times d_{\mathrm{model}}}$.

\subsubsection{Decoder}
A feedforward layer with sigmoid activation decodes the embeddings and predicts the probability for each point $emb_{i}$ to be part of a stay:
\begin{equation}
    \label{eq:method/sp-class/embeddings/prediction}
    \hat{c_{i}} = \sigma(emb_{i}~{W_{\mathrm{d}}}^T+b_{\mathrm{d}}),\quad~\hat{c}\in{[0,1]}^{n}
\end{equation}
Now the segmentation can be done by simply grouping consecutive points where $\hat{c_{i}}<0.5$ for non-SRs and $\hat{c_{i}}>0.5$ for SRs, respectively.

\subsubsection{Supervision}
\label{sec:method/sp-class/supervision}
In the case of available SR labels, we can compute the pointwise binary cross entropy (BCE) between the prediction $\hat{c}$ and the ground truth $c$:
\begin{equation}
    \mathrm{BCE}(\hat{c}_{i}, c_{i}) = -c_{i}\log \hat{c}_{i} - (1-c_{i})\log (1-\hat{c}_{i})
\end{equation}
The distribution of the binary labels can be highly imbalanced. To prevent the model leaning towards one of the classes, we apply class weighting based on the mean $\overline{c}_{\mathrm{train}}$ within the training set:
\begin{equation}
    \label{eq:method/sp-class/supervision/weighted-bce}
    \mathrm{BCE}_{\mathrm{w}}(\hat{c}_{i}, c_{i}, \overline{c}_{\mathrm{train}}) = \left(\frac{c_{i}}{\overline{c}_{\mathrm{train}}}+ \frac{1-c_{i}}{1-\overline{c}_{\mathrm{train}}}\right) \mathrm{BCE}(\hat{c}_{i}, c_{i}) 
\end{equation}
Now the total loss $\mathcal{L}_{\mathrm{super}}$ is the average weighted BCE over all points in all $N_{\mathrm{train}}$ training sequences:
\begin{equation}
    \label{eq:method/sp-class/supervision/total-loss}
    \mathcal{L}_{\mathrm{super}} = \frac{1}{N_{\mathrm{train}}\cdot n}\sum_{j=1}^{N_{\mathrm{train}}}{\sum_{i=1}^n{\mathrm{BCE}_{\mathrm{w}}(\hat{c}^{(j)}_{i}, c^{(j)}_{i},\overline{c}_{\mathrm{train}})}}
\end{equation}

\subsection{Weakly Supervised SR Extraction}
\label{sec:method/weak-super}%
Since the vast amount of publicly available location trajectories does not contain SR labels, we apply \emph{programmatic} weak supervision \cite{ratner2016data} by generating weak labels based on other data sources. These labels are often inaccurate. However, since we can generate them on a large scale, and since the error generally does not correlate with the input, we expect our model to still learn useful latent representations.

For that, we define a function $f_{\mathrm{weak}}$ that returns the estimated probability $c_{i_{\mathrm{weak}}}$ that the location point $g_{i}$ is part of a stay, and a confidence score $w_{i_{\mathrm{weak}}}$ for that prediction:
\begin{equation}
    f_{\mathrm{weak}}(g_{i}) = (c_{i_{\mathrm{weak}}}, w_{i_{\mathrm{weak}}})
\end{equation}
Here, $c_{i_{\mathrm{weak}}}$ replaces the ground truth value $c_{i}$, while $w_{i_{\mathrm{weak}}}$ is used to weight the influence of the weak label on the total loss. Thus, the model learns more from weak labels, where the labeling function is more certain. The total loss is then:
\begin{equation}
    \mathcal{L}_{\mathrm{weak}} = \sum_{j=1}^{N_{\mathrm{train}}}{\sum_{i=1}^n{\frac{w^{(j)}_{i_{\mathrm{weak}}}}{N_{\mathrm{train}}\cdot n}\mathrm{BCE}_{\mathrm{w}}(\hat{c}^{(j)}_{i}, c^{(j)}_{i_{\mathrm{weak}}},\overline{c}_{\mathrm{train_{weak}}})}}
    \label{eq:method/weak-super/loss}
\end{equation}
Furthermore, the mean label $\overline{c}_{\mathrm{train_{weak}}}$ is also weighted by the confidence score:
\begin{equation*}
    \overline{c}_{\mathrm{train_{weak}}} = \frac{\sum_{j=1}^{N_{\mathrm{train}}}{\sum_{i=1}^n{c^{(j)}_{i_{\mathrm{weak}}}\cdot w^{(j)}_{i_{\mathrm{weak}}}}}}{\sum_{j=1}^{N_{\mathrm{train}}}{\sum_{i=1}^n{w^{(j)}_{i_{\mathrm{weak}}}}}}
\end{equation*}

$f_{\mathrm{weak}}$ works as an ensemble of separate labeling functions that predict whether a location point is part of a stay or not. All labeling functions implement simple heuristics and may conflict with each other. Here, they predict a pair $(c_{\mathrm{weak}},w_{\mathrm{weak}}(g))$ with a constant value for $c_{\mathrm{weak}}$ and a confidence weight $w_{\mathrm{weak}}$ depending on the input data $g$. In total, four different functions are defined:
\begin{itemize}
    \item $f_{\mathrm{build}}$ predicts a stay with high confidence if a location lies within a building.
    \item $f_{\mathrm{am}}$ predicts a stay with high confidence if a location lies within small amenities.
    \item $f_{\mathrm{street}}$ predicts a non-stay with high confidence if a location is close to a street.
    \item $f_{\mathrm{transport}}$ predicts a non-stay based on available transportation mode labels.
\end{itemize}

The data source for the first three functions is OSM. Similar to \cite{cao2010mining}, the coordinates of the location $g_{i}$ are used to query additional information from the map service. 

\subsubsection{Stay Labeling Functions}

$f_{\mathrm{build}}$ checks, if $g_{i}$ lies within a building $b\in \mathcal{B}_{\mathrm{OSM}}$. If so, it returns a confidence weight of 1, since points that fall inside a building have a high chance of being part of a stay:

\begin{equation}
        w_{\mathrm{build}}(g_{i}) =
        \begin{cases}
            1, & \text{if}~~\exists~b\in \mathcal{B}_{\mathrm{OSM}}~~|~~b \cap g_{i}\neq \{\}\\
            0, & \text{otherwise}
        \end{cases}
\end{equation}

Similarly, $f_{\mathrm{am}}$ returns $w_{\mathrm{am}} > 0$, if $g_{i}$ lies within an amenity $a\in \mathcal{A}_{\mathrm{OSM}}$. This is an OSM category for facilities like hospitals or airports that can encapsulate multiple buildings. For larger amenities, since it is less certain that people will stay in a single location, we model the confidence weight as a function of their geographic area:

\begin{equation}
    \resizebox{.95\hsize}{!}{$
        w_{\mathrm{am}}(g_{i}) =
        \begin{cases} \max\limits_{a\in \mathcal{A}_{\mathrm{OSM}}|a\cap g_{i}\neq \{\}} \exp\left(-\frac{area(a)}{\frac{1}{|\mathcal{A}_{\mathrm{OSM}}|}\sum_j area(A_{{\mathrm{OSM}_j}})}\right) & \text{if}~\exists a|a\cap g_{i}\neq \{\}\\
            0, & \text{otherwise}
        \end{cases}
    $}
\end{equation}
Here, the fraction is the ratio of the area of an encapsulating amenity to the average area of all amenities in the dataset. By using a negative exponent, the weight starts at 1 for an area of 0 and decreases as the area increases. We use the maximum value, when multiple amenities encapsulate $g_{i}$.

\subsubsection{Non-Stay Labeling Functions}

$f_{\mathrm{street}}$ checks if $g_{i}$ is near a street, since those points have a high probability of being part of a non-stay. Thus, if a street $s\in \mathcal{S}_{\mathrm{OSM}}$ intersects with a centered bounding box $bb_{i}(l_s)$ around $g_{i}$, $f_{\mathrm{street}}$ returns a confidence weight of 1. The box has a shape of $d(l_s)\times d(l_s)$ with $l_s$ denoting the importance level of the street $s$ (e.g. highway). Formally, this can be defined as:

\begin{equation}
    w_{\mathrm{street}}(g_{i}) =
    \begin{cases}
        1, & \text{if}~~\exists~s\in \mathcal{S}_{\mathrm{OSM}} | s \cap bb_{i}(l_s) \neq \{\}\\
        0, & \text{otherwise}
    \end{cases}
\end{equation}

$f_{\mathrm{transport}}$ is designed for the GeoLife (GL) dataset, which will be introduced in Chapter \ref{sec:data} in more detail. Its trajectories include additional transportation mode labels, which are: walking, running, biking, motorcycle, car, taxi, bus, train, subway, boat, and airplane. While the dataset lacks a separate stay mode, we can use all motorized modes (i.e., all modes except walking, running, and biking) as a heuristic for non-stays. The confidence weight is formalized as:

\begin{equation}
        w_{\mathrm{transport}}(g_{i}) =
        \begin{cases}
            1, & \text{if}~label(g_{i})\in modes_{\mathrm{motorized}}\\
            0, & \text{otherwise}
        \end{cases}
\end{equation}

\subsubsection{Combining the Labeling Functions}
\label{sec:method/weak-super/combining}
We combine the results of all heuristics $\mathcal{H}$ by averaging the predicted probabilities and adding up all confidence weights as follows:

\begin{align}
    \begin{split}
        f_{\mathrm{weak}}(g_{i}) =& ~(c_{i_{\mathrm{weak}}}, w_{i_{\mathrm{weak}}})\\ =&
        \left(\frac{\sum_{j\in \mathcal{H}}c_{j}\cdot w_{j}(g_{i})}{\sum_{j\in \mathcal{H}}w_{j}(g_{i})},~~\sum_{j\in \mathcal{H}}w_{j}(g_{i})\right),
    \end{split}
\end{align}
\begin{equation*}
    \text{where }\mathcal{H} = \{build, am, street, transport\}.
\end{equation*}
Thus, each labeling function $f_{j}$ has a linear influence on the total confidence weight $w_{i_{\mathrm{weak}}}$, independent of the output of other labeling functions. On the one hand, this combination is similar to an ensemble with model averaging, whereas, on the other hand, this resembles also a Mixture of Experts, where the weights depend on the input $g_{i}$.

\subsection{Self-Supervised Encoder}
\label{sec:method/ssl}
Many points are not captured by any heuristic and receive a total confidence weight of 0. Self-supervised learning (SSL) could still leverage those data in a (weakly) semi-supervised manner and further strengthen the model's robustness to inaccurate training data \cite{hendrycks2019using}. Since \cite{jawed2020self,tipirneni2022self} show good results by using forecasting as a pretext task for time series data, we adopted their approach.

\subsubsection{Forecasting Task}
We choose the velocity as one forecast target, which is less dependent on the sample rate compared to the location. Additionally, the bearing angle is forecasted as a second target because it is not directly included in the input and requires the model to encode more informative embeddings. More specifically, we predict the sine and cosine values of the angle to capture the periodicity.

Given the encoder output $emb$, we concatenate its sequence mean $\overline{emb}$ and last embedding vector $emb_{n}$ as an aggregated embedding vector $emb_{\mathrm{agg}}\in\mathbb{R}^{2d_{\mathrm{model}}}$ for the whole sequence. This vector is then passed to two separate feedforward layers. No activation function is used for the velocity, while for the sine and cosine prediction we apply tanh to bind the output between -1 and 1:
\begin{equation}
    \hat{v}_{n+1} = emb_{\mathrm{agg}}~{W_{\mathrm{vel}}}^T+b_{\mathrm{vel}}
\end{equation}
\begin{equation}
    \begin{bmatrix}\hat{sin}_{\alpha_{\mathrm{n+1}}} \\ \hat{cos}_{\alpha_{\mathrm{n+1}}}  \end{bmatrix} = \tanh\left(emb_{\mathrm{agg}}~{W_{\mathrm{ang}}}^T+b_{\mathrm{ang}}\right)
\end{equation}

\subsubsection{Multitask Loss}
The loss for each pretext task is defined by the Mean Squared Error (MSE) between the prediction and the ground truth:
\begin{equation}
    \mathrm{MSE}(\hat{y}, y) = \frac{1}{N_{\mathrm{train}}}\sum_{j=1}^{N_{\mathrm{train}}}{\|\hat{y}^{(j)}_{n+1} - y^{(j)}_{n+1}\|}
\end{equation}
\begin{equation}
    \mathcal{L}_{\mathrm{vel}} = \mathrm{MSE}(\hat{v}, v)
\end{equation}
\begin{equation}
    \mathcal{L}_{\mathrm{ang}} = \mathrm{MSE}(\hat{sin}_{\alpha}, \sin\alpha) + \mathrm{MSE}(\hat{cos}_{\alpha}, \cos\alpha)
\end{equation}

We follow \cite{jawed2020self} and approach SSL as multitask learning with the sum of the downstream loss $\mathcal{L}_{\mathrm{weak}}$ and the pretext losses:
\begin{equation}
    \label{eq:method/ssl/final-loss}
    \mathcal{L}_{\mathrm{final}} = \mathcal{L}_{\mathrm{weak}} + \lambda_{\mathrm{vel}}\mathcal{L}_{\mathrm{vel}} + \lambda_{\mathrm{ang}}\mathcal{L}_{\mathrm{ang}} ,
\end{equation}
where $\lambda$ denotes tunable hyperparameters. If ground truth labels are available, $\mathcal{L}_{\mathrm{weak}}$ is replaced with $\mathcal{L}_{\mathrm{super}}$.

\section{Data}
\label{sec:data}
For this study, we select two datasets: GeoLife (GL) by \citegeolife\ and ExtraSensory (ES) by \cite{vaizman2017recognizing}. GL contains two orders of magnitude more location points than ES but lacks proper SR labels. ES is chosen because of its activity labels, from which we can infer ground truth SR labels.

Similar to \cite{dabiri2019semi} and \cite{zheng2008learning}, we remove outliers based on unrealistic velocity values and split a user's trajectory if the time difference between two consecutive points exceeds 20 minutes or if an unrealistic location jump is detected.

\subsection{GeoLife}
\label{sec:exp/data/gl}
Instead of ground truth SR labels, the GL dataset contain time-segmented transportation mode labels from 69 of all 182 participants. These labels are used to derive weak labels and for our experiment on TMD.

To reduce network traffic and memory, we gather OSM data for points that fall within the $15 \%$/$85 \%$ percentile of longitude and latitude, which covers about $62 \%$ of the total dataset. An overview of the total sum of confidence weights used for weak supervision can be found in Table \ref{tab:exp/data/gl}. The remaining unlabeled data is still used for SSL instead. The sample rate of GL is non-constant and varies between 1 and 6 seconds. For the UTM projection, we choose zone 50N.

\begin{table}[]
    \centering
    \caption{Summary of weak labels derived from GL dataset.}
    \begin{tabular}{llc}
    \hline
    \multicolumn{1}{c}{\textbf{weak label}} & \multicolumn{1}{c}{\textbf{heuristic}} & \multicolumn{1}{c}{\textbf{\begin{tabular}[c]{@{}c@{}}total sum of\\ confidence weights\end{tabular}}} \\ \hline
    \multirow{2}{*}{stays ($c_{\mathrm{weak}}=1$)} & building & 1.0 M\\
    & amenity & 0.2 M\\\hline
    \multirow{2}{*}{non-stays ($c_{\mathrm{weak}}=0$)} & street & 5.0 M\\
    & transport & 2.7 M\\\hline
    \end{tabular}
    \label{tab:exp/data/gl}
\end{table}

\subsection{ExtraSensory}
\label{sec:exp/data/es}
We use the ES dataset to fine-tune and evaluate DeepStay. Besides GNSS points, this dataset contains other sensor data, which we ignore. It was collected for the task of activity recognition. Participants should self-report their current activities such as "biking" or "watching TV". Some activity modes clearly indicate stays and non-stays. Thus, we define a function that maps these modes to SR labels.

In the second step, we remove suspicious stays, where the velocity is higher than the average velocity of non-stays. The final number of points and derived labels are listed in Table \ref{tab:exp/data/es}.
\begin{table}[]
    \centering
    \caption{Summary of the cleaned ES dataset.}
    \begin{tabular}{lc}
    \hline
    \multicolumn{1}{c}{\textbf{}} & \multicolumn{1}{c}{\textbf{total number}} \\ \hline
    GNSS points & 306 k \\ 
    stays ($c_i=1$) & 223 k \\ 
    non-stays ($c_i=0$)  & 28 k\\\hline
    \end{tabular}
    \label{tab:exp/data/es}
\end{table}
The sample rate of ES is nearly constant at $\frac{1}{\mathrm{min}}$. To achieve reasonable results with an encoder pre-trained on GL, we linearly interpolate the location trajectory at a rate of $\SI{0.5}{Hz}$. However, for the final test results, only the predictions for the real, non-interpolated labels are evaluated. The prediction value is taken from the nearest interpolation point. For the map projection, we use the UTM zone 11N.

\section{Experiments}
\label{sec:exp}
In the first experiment, we train and test DeepStay on SR labels. The second experiment shows the ability of our architecture to be used for the more general task of TMD.

\subsection{Experiment 1: Stay Region Extraction}
\label{sec:exp/stay}
For this experiment, DeepStay is pre-trained on weak labels from the GL dataset and then fine-tuned and tested together with traditional baselines on the ES dataset, where it achieves the best overall results among all methods.

\subsubsection{Baselines}
\label{sec:exp/stay/baselines}
We implement the following algorithms as baselines and test them on the ES dataset:

\begin{itemize}
    \item \textbf{Kang et al. \cite{kang2005extracting}}: Threshold-based clustering. It collects consecutive points until a distance threshold to the points' centroid is exceeded. Then the time criterion \ref{eq:t_tresh} is checked and if the minimum duration is reached, the collected points form a SR. Although the authors only proposed a POI extraction algorithm, it also implicitly incorporates SR extraction, which can be outsourced.
    \item \textbf{D-Star \cite{nishida2015extracting}}: Density-based clustering. It is based on DBSCAN, but instead of solely clustering the location points spatially, it considers only neighboring points along the trajectory and tries to exclude outliers. D-Star seems to be state-of-the-art.
    \item \textbf{CB-SMoT \cite{cbsmot}}: Density-based clustering. While the algorithm is similar to D-Star, the resulting SRs contain only consecutive points, which is more in line with our definition. It can incorporate prior known POIs. However, for a fair comparison, we exclude this data.
\end{itemize}
We optimize the hyperparameters of Kang et al. and CB-SMoT using a $3\times3$ grid search based on the values reported in the original publications. D-Star has 4 parameters to adjust, hence we perform a random search with 10 different constellations. Each parameter search is incorporated in a 5-fold cross-validation based on the $F_1$ score. We split the ES data in the same way as for DeepStay.

\subsubsection{Training, Validation, and Test}
\label{sec:exp/stay/train-val}
The training and testing pipeline for DeepStay can be summarized in three steps:
\begin{enumerate}
    \item \textbf{Hyperparameter optimization}: Training on about 80 \% of the GL dataset with weak labels and optimization of hyperparameters on the remaining $\SI{20}{\%}$ in respect to the loss $\mathcal{L}_{\mathrm{weak}}$. These hyperparameters are the number of training epochs, the weight decay, the learning rate, and the SSL weights $\lambda_{\mathrm{vel}}$ and $\lambda_{\mathrm{ang}}$.
    \item \textbf{Pre-training}: Creating a pre-trained DeepStay model by reinitiating the training on the full GL dataset and using the best-known hyperparameters.
    \item \textbf{Fine-tuning and test}: Fine-tuning the decoder of the pre-trained model on the ES dataset and freezing all other model weights including the encoder layers. We apply 5-fold cross-validation, i.e. each iteration about 80 \% of the data is used for training, and validation and 20 \% for testing. Of this 80 \%, 10 \% is used for a second hyperparameter optimization.
\end{enumerate}
We follow \cite{etemad2018transportation} and split the data by the participants of the respective study, to avoid leakage between training, validation and test set. During both the pre-training and the fine-tuning, we apply an Adam optimizer \cite{kingma2014adam} and SSL.

\subsubsection{Metrics}
\label{sec:exp/stay/metrics}
A common metric in time series segmentation is the pointwise accuracy, i.e. the ratio of correctly classified points to the total number of labels.

In addition, we measure the pointwise calculated recall and precision. The definition of the positive class is crucial for both metrics. Since the final test dataset, i.e. ES, is highly imbalanced and contains many more stays than non-stays (see Table \ref{tab:exp/data/es}), it is more important to detect a non-stay than a stay. This also resembles everyday life, where people mostly stay in one place and only move from time to time.  Therefore, we choose non-stays as the positive class. The derived $F_1$ score is used as the main metric to evaluate all SR extraction algorithms.

\subsubsection{Results}
\label{sec:exp/stay/results}
\begin{table}[]
    \centering
    \caption{Final results on the ES dataset.}
    \begin{tabular}{lccccc}
    \hline
    \textbf{Method} & \textbf{$F_1$} & \textbf{Acc.} & \textbf{Precision} & \textbf{Recall}\\
\hline
DeepStay (ours) & \textbf{0.788} & \textbf{0.954} & 0.822 & 0.757 \\
\hline
D-Star \cite{nishida2015extracting}       & 0.753 & 0.951 & \textbf{0.877} & 0.660 \\
CB-SMoT  \cite{cbsmot}     & 0.548 & 0.909 & 0.619 & 0.491 \\
Kang et al.  \cite{kang2005extracting}  & 0.453 & 0.796 & 0.325 & 0.748 \\
\hline
constant $\hat{c}_i=0$ & 0.203 & 0.113 & 0.113 & \textbf{1.000} \\
constant $\hat{c}_i=1$ & 0.000 & 0.887 & -     & 0.000 \\
\hline
    \end{tabular}
    \label{tab:exp/stay/results}
\end{table}

The final results are shown in Table \ref{tab:exp/stay/results}. All reported values are calculated over all 5 ES test data splits. In addition to the three baselines, two simplistic baselines predict a constant value (either always non-stay $\hat{c}_i=0$ or always stay $\hat{c}_i=1$). 

DeepStay achieves higher overall scores than all implemented baselines, while the results for D-Star are comparable in terms of accuracy. Kang et al. use an approach with hard thresholds, which seems to be disadvantageous compared to a density-based approach. Even though CB-SMoT achieves relatively high accuracy, its $F_1$ score is significantly worse than the similar D-Star algorithm. This may be due to the missing outlier detection in CB-SMoT.

\subsubsection{Ablation Study}
\label{sec:exp/stay/ablation}

We compare the contribution of different training components in Table \ref{tab:exp/stay/ablation}, where we analyze the effect of training DeepStay first without any SSL and second without any pre-training, i.e. solely trained on the ES dataset. For the latter, the original sample rate of $\frac{1}{\mathrm{min}}$ was used instead of interpolation. In addition to the previous metrics, we also measure the area under the PR curve (PR-AUC). It can be seen that the effect of SSL is relatively small. However, the pre-training has a significant impact on the performance, showing that the model correctly handles the noise coming from the weak labels and learns reasonable latent representations of the SRs.

\begin{table}[]
    \centering
    \caption{Ablation study of DeepStay tested on ES.}
    \begin{tabular}{lccccc}
    \hline
    \textbf{Method} & \textbf{$F_1$} & \textbf{Acc.} & \textbf{PR-AUC}
    & \textbf{Precision} & \textbf{Recall}\\
\hline
DeepStay (full)  & \textbf{0.788} & \textbf{0.954} & \textbf{0.821} & 0.822 & 0.757 \\
 w/o SSL      & 0.780 & 0.953 & 0.809 & \textbf{0.837} & 0.729 \\
 w/o pre-training    & 0.557 & 0.850 & 0.787 & 0.418 & \textbf{0.838} \\
\hline
    \end{tabular}
    \label{tab:exp/stay/ablation}
\end{table}

\subsection{Experiment 2: Transportation Mode Detection}
\label{sec:exp/transp}
To further demonstrate the broader applicability of DeepStay and to contribute our findings to a broader field of research, we apply the same encoder for TMD.

There has been some work on transformer-based TMD for data other than trajectories, such as accelerometer, gyroscope, and magnetometer data \cite{tian2022transportation}. However, these sensors are sampled at a much higher rate ($>\SI{20}{Hz}$) and thus the input sequences cover only a few seconds. In this case, the transportation mode is mostly constant, so the segmentation part is dropped from the TMD task and only the classification part remains.

For TMD \emph{from location trajectories}, sequences typically cover several minutes and therefore mode changes are likely to occur. Nevertheless, most of the related work presupposes a correct segmentation and simply classifies each of the segments as one of the available modes \cite{moreau2021devil}. This is problematic for real-world applications, where a correct segmentation is never given in advance. Here, the advantage of using the transformer encoder is the joint segmentation and classification of transportation modes by simply predicting the pointwise class probabilities and grouping consecutive predicted points of the same modes together. The baseline model SECA \cite{dabiri2019semi} may be the state-of-the-art approach of those models that segment \emph{and} classify from raw GNSS trajectories. Although their published code lacks the segmentation part, we compare their self-reported results on the GL dataset with our own results and show that our approach significantly outperforms SECA.

\subsubsection{Model Adaptations}
The only adaptations made to DeepStay are in the decoder and the supervision. The decoder's weights are expanded and a softmax activation predicts the pointwise probability $\hat{c}_{i,m}$ for each of the $M=5$ transportation modes:
\begin{equation}
    \hat{c}_{i,m} = \mathrm{softmax}(emb_{i}~{W_{\mathrm{d^\prime}}}^T+b_{\mathrm{d^\prime}})_m,\quad~\hat{c}\in{[0,1]}^{n\times M}
\end{equation}
Now $\mathrm{BCE}_{\mathrm{w}}$ in \ref{eq:method/sp-class/supervision/total-loss} is replaced by the weighted cross entropy ($\mathrm{CE}_{\mathrm{w}}$) in \ref{eq:exp/transp/ce} between prediction and ground truth $c_{i,m}$, where $\overline{c}_{\mathrm{train}_m}$ denotes the percentage of labels of the $m$-th class within the training set. For segmentation, we can simply group consecutive points with the same most probable class.
\begin{equation}
    \label{eq:exp/transp/ce}
    \mathrm{CE}_{\mathrm{w}}(\hat{c}_{i},c_{i}, \overline{c}_{\mathrm{train}}) = -\sum_{m=1}^{M} c_{i,m}\frac{\log(\hat{c}_{i,m})}{M\cdot\overline{c}_{\mathrm{train}_m}}
\end{equation}

\subsubsection{Baseline and Comparable Datasets}
The SECA model \cite{dabiri2019semi} is used as the only baseline. The authors perform a change point search by using the PELT method \cite{killick2012optimal} to first segment the trajectory. Second, they use a convolutional neural network (CNN) to predict the mode of each segment and integrate an autoencoder for semi-supervision.

We compare the size of the dataset after our own preprocessing with that of SECA in Table \ref{tab:exp/transp/datasets}. It shows that we train DeepStay with significantly fewer labels compared to SECA. However, in total more unlabeled data is available. Overall, the test sets are quite similar, which allows us to compare the final results of DeepStay and SECA.
\begin{table}[]
    \centering
    \caption{Total number of GNSS points after preprocessing.}
    \begin{tabular}{lccccccc}
    \hline
     & \textbf{\begin{tabular}[c]{@{}c@{}}unlabeled\\ (Training)\end{tabular}} & \textbf{\begin{tabular}[c]{@{}c@{}}labeled\\ (Training)\end{tabular}} & \textbf{\begin{tabular}[c]{@{}c@{}}labeled\\ (Test)\end{tabular}} \\
\hline

This work                  & 16.29 M & 3.74 M & 4.76 M\\
SECA \cite{dabiri2019semi} & 15.43 M & 4.76 M & 4.76 M\\
\hline
    \end{tabular}
    \label{tab:exp/transp/datasets}
\end{table}

\subsubsection{Training, Validation, and Test}
\label{sec:exp/transp/train-val}
Both SECA and DeepStay are trained semi-supervised. While SECA uses an autoencoder, DeepStay applies SSL (see Section \ref{sec:method/ssl}).

Unlike in Experiment 1, we randomly assign each sequence $seq$ to one of the training or test sets, regardless of the including participants, to match the setup of SECA. We also apply 5-fold cross-validation. In addition, 20\% of the training data is used to adjust the same hyperparameters as in Experiment 1. We optimize our model using Adam \cite{kingma2014adam}.

\subsubsection{Results}
\label{sec:exp/transp/results}
We report the weighted $F_1$ score and the accuracy. This $F_1$ score is the average of the per-class $F_1$ scores weighted by the number of labels per class. SECA is performing segmentwise classification and DeepStay pointwise classification, thus the following results are not fully comparable.

\begin{table}[]
    \centering
    \caption{Final results for TMD tested on the GL dataset.}
    \begin{tabular}{lcc}
    \hline
    \textbf{Method} & \textbf{$F_1$} & \textbf{Acc.}\\
\hline
		
DeepStay (ours)  & \textbf{0.830} & \textbf{0.831}\\
SECA with ground truth segments & 0.764 & 0.768\\
SECA with predicted segments    & 0.717 & 0.721\\
\hline
    \end{tabular}
    \label{tab:exp/transp/results}
\end{table}

Nevertheless, the final results in Table \ref{tab:exp/transp/results} clearly demonstrate the significant performance improvement of DeepStay. A major reason may be the pointwise predictions, which do not require a prior segmentation, but intrinsically segment the data for the classification task. However, even when SECA is given ground truth segments, DeepStay still achieves better results. One reason may be that, unlike SECA, the input sequence for our model is not limited to a single transportation mode, i.e., it can also learn the transition between modes. E.g., it is intuitively more likely to see a transition from bus to train than from bus to car. Furthermore, the autoencoder in SECA only tries to reconstruct the trajectory, while SSL can provide proxy labels for DeepStay, which may be more informative. In addition, the transformer model with its attention mechanism seems to be superior in comparison to the CNN layers for this task.

\section{Conclusion and Future Work}
In this work, we show, how to derive programmatically weak labels for SR extraction and how to successfully train a transformer encoder with these data. We demonstrate the effectiveness of this model on ground truth data for SR extraction and TMD, where it outperforms state-of-the-art methods. This work should be seen as a starting point for new data-driven approaches to SR extraction and provides useful training and test data. Ideas for future work are:
\subsubsection{More data augmentation}
Instead of always training on the same sequences, all trajectories could be shifted by a number of points in each epoch. This results in slightly different sequences and SSL targets and reduces overfitting.
\subsubsection{Modeling dependencies}
We treat all labeling functions independently, although there are clear dependencies. E.g., $w_{\mathrm{build}}$ correlates strongly with $w_{\mathrm{am}}$, because buildings are often part of amenities. Other work suggests that the performance benefits significantly from incorporating these dependencies \cite{NIPS2017_cedebb6e}.
\subsubsection{Pre-training on multiple datasets}
In this study, we stick with the GL dataset for pre-training. However, there are many public unlabeled GNSS trajectory datasets. All of them could be weakly labeled with our approach and carefully combined to have an even larger training set.


\addtolength{\textheight}{-12cm}   





\bibliography{./biblio}

\begin{thebibliography}{10}
\providecommand{\url}[1]{#1}
\csname url@samestyle\endcsname
\providecommand{\newblock}{\relax}
\providecommand{\bibinfo}[2]{#2}
\providecommand{\BIBentrySTDinterwordspacing}{\spaceskip=0pt\relax}
\providecommand{\BIBentryALTinterwordstretchfactor}{4}
\providecommand{\BIBentryALTinterwordspacing}{\spaceskip=\fontdimen2\font plus
\BIBentryALTinterwordstretchfactor\fontdimen3\font minus
  \fontdimen4\font\relax}
\providecommand{\BIBforeignlanguage}[2]{{%
\expandafter\ifx\csname l@#1\endcsname\relax
\typeout{** WARNING: IEEEtran.bst: No hyphenation pattern has been}%
\typeout{** loaded for the language `#1'. Using the pattern for}%
\typeout{** the default language instead.}%
\else
\language=\csname l@#1\endcsname
\fi
#2}}
\providecommand{\BIBdecl}{\relax}
\BIBdecl

\bibitem{vaswani2017attention}
A.~Vaswani, N.~Shazeer, N.~Parmar, J.~Uszkoreit, L.~Jones, A.~N. Gomez,
  {\L}.~Kaiser, and I.~Polosukhin, ``Attention is all you need,''
  \emph{Advances in neural information processing systems}, vol.~30, 2017.

\bibitem{zheng2010understanding}
Y.~Zheng, Y.~Chen, Q.~Li, X.~Xie, and W.-Y. Ma, ``Understanding transportation
  modes based on gps data for web applications,'' \emph{ACM Transactions on the
  Web (TWEB)}, vol.~4, no.~1, pp. 1--36, 2010.

\bibitem{suzuki2019personalized}
J.~Suzuki, Y.~Suhara, H.~Toda, and K.~Nishida, ``Personalized visited-poi
  assignment to individual raw gps trajectories,'' \emph{ACM Transactions on
  Spatial Algorithms and Systems (TSAS)}, vol.~5, no.~3, pp. 1--28, 2019.

\bibitem{kang2005extracting}
J.~H. Kang, W.~Welbourne, B.~Stewart, and G.~Borriello, ``Extracting places
  from traces of locations,'' \emph{ACM SIGMOBILE Mobile Computing and
  Communications Review}, vol.~9, no.~3, pp. 58--68, 2005.

\bibitem{li2008mining}
Q.~Li, Y.~Zheng, X.~Xie, Y.~Chen, W.~Liu, and W.-Y. Ma, ``Mining user
  similarity based on location history,'' in \emph{Proceedings of the 16th ACM
  SIGSPATIAL international conference on Advances in geographic information
  systems}, 2008, pp. 1--10.

\bibitem{zheng2009mining}
Y.~Zheng, L.~Zhang, X.~Xie, and W.-Y. Ma, ``Mining interesting locations and
  travel sequences from gps trajectories,'' in \emph{Proceedings of the 18th
  international conference on World wide web}, 2009, pp. 791--800.

\bibitem{andrade2020discovering}
T.~Andrade, B.~Cancela, and J.~Gama, ``Discovering locations and habits from
  human mobility data,'' \emph{Annals of Telecommunications}, vol.~75, no.~9,
  pp. 505--521, 2020.

\bibitem{umair2014discovering}
M.~Umair, W.~S. Kim, B.~C. Choi, and S.~Y. Jung, ``Discovering personal places
  from location traces,'' in \emph{16th International Conference on Advanced
  Communication Technology}.\hskip 1em plus 0.5em minus 0.4em\relax IEEE, 2014,
  pp. 709--713.

\bibitem{bhattacharya2012extracting}
T.~Bhattacharya, L.~Kulik, and J.~Bailey, ``Extracting significant places from
  mobile user gps trajectories: a bearing change based approach,'' in
  \emph{Proceedings of the 20th International Conference on Advances in
  Geographic Information Systems}, 2012, pp. 398--401.

\bibitem{pavan2015finding}
M.~Pavan, S.~Mizzaro, I.~Scagnetto, and A.~Beggiato, ``Finding important
  locations: A feature-based approach,'' in \emph{2015 16th IEEE International
  Conference on Mobile Data Management}, vol.~1.\hskip 1em plus 0.5em minus
  0.4em\relax IEEE, 2015, pp. 110--115.

\bibitem{thomason2016identifying}
A.~Thomason, N.~Griffiths, and V.~Sanchez, ``Identifying locations from
  geospatial trajectories,'' \emph{Journal of Computer and System Sciences},
  vol.~82, no.~4, pp. 566--581, 2016.

\bibitem{zheng2008understanding}
Y.~Zheng, Q.~Li, Y.~Chen, X.~Xie, and W.-Y. Ma, ``Understanding mobility based
  on gps data,'' in \emph{Proceedings of the 10th international conference on
  Ubiquitous computing}, 2008, pp. 312--321.

\bibitem{pavan2017mining}
M.~Pavan, S.~Mizzaro, and I.~Scagnetto, ``Mining movement data to extract
  personal points of interest: A feature based approach,'' in \emph{Information
  Filtering and Retrieval}.\hskip 1em plus 0.5em minus 0.4em\relax Springer,
  2017, pp. 35--61.

\bibitem{ester1996density}
M.~Ester, H.-P. Kriegel, J.~Sander, X.~Xu \emph{et~al.}, ``A density-based
  algorithm for discovering clusters in large spatial databases with noise.''
  in \emph{kdd}, vol.~96, no.~34, 1996, pp. 226--231.

\bibitem{ankerst1999optics}
M.~Ankerst, M.~M. Breunig, H.-P. Kriegel, and J.~Sander, ``Optics: Ordering
  points to identify the clustering structure,'' \emph{ACM Sigmod record},
  vol.~28, no.~2, pp. 49--60, 1999.

\bibitem{cbsmot}
A.~T. Palma, V.~Bogorny, B.~Kuijpers, and L.~O. Alvares, ``A clustering-based
  approach for discovering interesting places in trajectories,'' in
  \emph{Proceedings of the 2008 ACM symposium on Applied computing}, 2008, pp.
  863--868.

\bibitem{nishida2015extracting}
K.~Nishida, H.~Toda, and Y.~Koike, ``Extracting arbitrary-shaped stay regions
  from geospatial trajectories with outliers and missing points,'' in
  \emph{Proceedings of the 8th ACM SIGSPATIAL International Workshop on
  Computational Transportation Science}, 2015, pp. 1--6.

\bibitem{ashbrook2003using}
D.~Ashbrook and T.~Starner, ``Using gps to learn significant locations and
  predict movement across multiple users,'' \emph{Personal and Ubiquitous
  computing}, vol.~7, no.~5, pp. 275--286, 2003.

\bibitem{cao2010mining}
X.~Cao, G.~Cong, and C.~S. Jensen, ``Mining significant semantic locations from
  gps data,'' \emph{Proceedings of the VLDB Endowment}, vol.~3, no. 1-2, pp.
  1009--1020, 2010.

\bibitem{stylianou2017stay}
G.~Stylianou, ``Stay-point identification as curve extrema,'' \emph{arXiv
  preprint arXiv:1701.06276}, 2017.

\bibitem{langley1998utm}
R.~B. Langley, ``The utm grid system,'' \emph{GPS world}, vol.~9, no.~2, pp.
  46--50, 1998.

\bibitem{moreau2021devil}
H.~Moreau, A.~Vassilev, and L.~Chen, ``The devil is in the details: An
  efficient convolutional neural network for transport mode detection,''
  \emph{IEEE Transactions on Intelligent Transportation Systems}, 2021.

\bibitem{ratner2016data}
A.~J. Ratner, C.~M. De~Sa, S.~Wu, D.~Selsam, and C.~R{\'e}, ``Data programming:
  Creating large training sets, quickly,'' \emph{Advances in neural information
  processing systems}, vol.~29, 2016.

\bibitem{hendrycks2019using}
D.~Hendrycks, M.~Mazeika, S.~Kadavath, and D.~Song, ``Using self-supervised
  learning can improve model robustness and uncertainty,'' \emph{Advances in
  Neural Information Processing Systems}, vol.~32, 2019.

\bibitem{jawed2020self}
S.~Jawed, J.~Grabocka, and L.~Schmidt-Thieme, ``Self-supervised learning for
  semi-supervised time series classification,'' in \emph{Pacific-Asia
  Conference on Knowledge Discovery and Data Mining}.\hskip 1em plus 0.5em
  minus 0.4em\relax Springer, 2020, pp. 499--511.

\bibitem{tipirneni2022self}
S.~Tipirneni and C.~K. Reddy, ``Self-supervised transformer for sparse and
  irregularly sampled multivariate clinical time-series,'' \emph{ACM
  Transactions on Knowledge Discovery from Data (TKDD)}, vol.~16, no.~6, pp.
  1--17, 2022.

\bibitem{zheng2010geolife}
Y.~Zheng, X.~Xie, W.-Y. Ma \emph{et~al.}, ``Geolife: A collaborative social
  networking service among user, location and trajectory.'' \emph{IEEE Data
  Eng. Bull.}, vol.~33, no.~2, pp. 32--39, 2010.

\bibitem{vaizman2017recognizing}
Y.~Vaizman, K.~Ellis, and G.~Lanckriet, ``Recognizing detailed human context in
  the wild from smartphones and smartwatches,'' \emph{IEEE pervasive
  computing}, vol.~16, no.~4, pp. 62--74, 2017.

\bibitem{dabiri2019semi}
S.~Dabiri, C.-T. Lu, K.~Heaslip, and C.~K. Reddy, ``Semi-supervised deep
  learning approach for transportation mode identification using gps trajectory
  data,'' \emph{IEEE Transactions on Knowledge and Data Engineering}, vol.~32,
  no.~5, pp. 1010--1023, 2019.

\bibitem{zheng2008learning}
Y.~Zheng, L.~Liu, L.~Wang, and X.~Xie, ``Learning transportation mode from raw
  gps data for geographic applications on the web,'' in \emph{Proceedings of
  the 17th international conference on World Wide Web}, 2008, pp. 247--256.

\bibitem{etemad2018transportation}
M.~Etemad, ``Transportation modes classification using feature engineering,''
  \emph{arXiv preprint arXiv:1807.10876}, 2018.

\bibitem{kingma2014adam}
D.~P. Kingma and J.~Ba, ``Adam: A method for stochastic optimization,''
  \emph{arXiv preprint arXiv:1412.6980}, 2014.

\bibitem{tian2022transportation}
Y.~Tian, D.~Hettiarachchi, and S.~Kamijo, ``Transportation mode detection
  combining cnn and vision transformer with sensors recalibration using
  smartphone built-in sensors,'' \emph{Sensors}, vol.~22, no.~17, p. 6453,
  2022.

\bibitem{killick2012optimal}
R.~Killick, P.~Fearnhead, and I.~A. Eckley, ``Optimal detection of changepoints
  with a linear computational cost,'' \emph{Journal of the American Statistical
  Association}, vol. 107, no. 500, pp. 1590--1598, 2012.

\bibitem{NIPS2017_cedebb6e}
P.~Varma, B.~D. He, P.~Bajaj, N.~Khandwala, I.~Banerjee, D.~Rubin, and
  C.~R\'{e}, ``Inferring generative model structure with static analysis,'' in
  \emph{Advances in Neural Information Processing Systems}, I.~Guyon, U.~V.
  Luxburg, S.~Bengio, H.~Wallach, R.~Fergus, S.~Vishwanathan, and R.~Garnett,
  Eds., vol.~30.\hskip 1em plus 0.5em minus 0.4em\relax Curran Associates,
  Inc., 2017.

\end{thebibliography}
\bibliographystyle{IEEEtran}

\end{document}